# Dialogue You Can Trust: Human and AI Perspectives on Generated Conversations

Ike Ebubechukwu[1,3,4], Johane Takeuchi[2], Antonello Ceravola[1], Frank Joublin[1] Marc Tanti[3]

August 2024

[1]Honda Research Institute Europe, Offenbach am Main, Germany
[2]Honda Research Institute Japan, Wako, Japan
[3]University of Malta, Malta
[4]University of Trento, Italy

**Abstract**

As dialogue systems and chatbots increasingly integrate into everyday interactions, the need for efficient and accurate evaluation methods becomes paramount. This study explores the comparative performance of human and AI assessments across a range of dialogue scenarios, focusing on seven key performance indicators (KPIs): Coherence, Innovation, Concreteness, Goal Contribution, Commonsense Contradiction, Incorrect Fact, and Redundancy. Utilizing the GPT-4o API, we generated a diverse dataset of conversations and conducted a two-part experimental analysis.

In Experiment 1, we evaluated multi-party conversations on Coherence, Innovation, Concreteness, and Goal Contribution, revealing that GPT models align closely with human judgments. Notably, both human and AI evaluators exhibited a tendency towards binary judgment rather than linear scaling, highlighting a shared challenge in these assessments.

Experiment 2 extended the work of Finch et al. (2023) by focusing on dyadic dialogues and assessing Commonsense Contradiction, Incorrect Fact, and Redundancy. The results indicate that while GPT-4o demonstrates strong performance in maintaining factual accuracy and commonsense reasoning, it still struggles with reducing redundancy and self-contradiction.

Our findings underscore the potential of GPT models to closely replicate human evaluation in dialogue systems, while also pointing to areas for improvement. This research offers valuable insights for advancing the development and implementation of more refined dialogue evaluation methodologies, contributing to the evolution of more effective and human-like AI communication tools.

## 1 Introduction

Communication has been a fundamental aspect of life since its origin, serving as the primary mechanism for interaction among living organisms. As the need to enhance the quality of these interactions emerged, a meta-level evaluation of communication effectiveness was introduced, either in real-time (during the dialogue) or post-session (based on notes or transcripts). Initially, basic evaluations relied on established resources, such as the Oxford and Cambridge dictionaries, to assess the contextual appropriateness of words and terms within dialogues [45]. With the advent of linguistics, more sophisticated metrics were developed to systematically evaluate dialogue interactions, addressing the growing need for comprehensive evaluation methods [14]. However, while human evaluators and annotators have proven effective, their involvement has been resource-intensive, costly, and susceptible to biases and errors [15]. However, to enhance efficiency and accuracy in text evaluation, tools like Microsoft Word were developed to assist human experts by automatically detecting spelling and grammatical errors in large volumes of text, thereby streamlining the review process and reducing the manual workload [23, 48].

With advancements in technology, several different models were developed to evaluate human dialogues more effectively by analyzing usage of words, sentence structures or even sentiment analysis [39]. Some of those models (particularly the one that could be automated) found applications on various social media platforms for tasks such as detecting racial slurs and other inappropriate content ([10], [50]). Despite their success, these systems were limited by their inability to fully replicate human understanding and nuance [4].

The introduction of artificial intelligence (AI) systems, particularly chatbots [52], revolutionized the way we engage in dialogues. These chatbots, trained on extensive datasets, are now used by approximately 80% of companies for customer interaction [19][1]. Human expertise combined with machine learning models began to evaluate these chatbot responses, ensuring accuracy and quality of the interactions [44]. However, these evaluations were primarily focused on surface-level metrics such as word accuracy and sentence comprehension [25].

Advancements in conversational AI have also enabled the creation of robots capable of engaging in both task-oriented and non-task-oriented dialogues, seamlessly switching between them based on context and speech recognition outcomes [37]. This capability is crucial for robots operating in dynamic environments where both functional and social interactions are essential. The integration of social dialogue has been shown to significantly enhance a robot's influence on human decision-making, highlighting the importance of combining relational and task-oriented dialogues in human-robot interactions [33].

The advent of Generative Pre-trained Transformers (GPT), particularly the black-box model GPT-3, significantly demonstrated a large improvements in the generation of human-like conversations [6]. Following the progress of GPT-3, several other models have been created by different institutions (establishes companies, university, startups), for instance models like LLaMA further advanced this field by showing performance comparable to leading model, although using less parameter or less training data [36]. These large language models (LLMs) generate conversations that are almost indistinguishable from those of humans. However, despite their impressive capabilities, users noted issues such as bias and hallucinations, which undermined trust in AI systems[4] [28]

---
[1]Netomi
[2]JPT

Despite these challenges, LLMs have become widely accepted and integrated into various sectors, from education to everyday smartphone use[40][2]. As these models became more sophisticated, systems using such model can now carry-out a conversation with or without human intervention [21]. This led to the rising needs for new approaches to dialogue evaluation, with the purpose of generating more reliable and trustworthy AI dialogues.

Our aim is to contribute in this area of automatic dialog evaluation through the usage of the newest model GPT-4o, comparing both human annotators and GPT-4o itself for evaluation. Our hypothesis posits that if GPT-4o can evaluate dialogues similarly to humans, two main key outcomes will emerge: 1) the reliance on expensive human annotators will decrease, and 2) the opportunity of using Multi-Agent Systems for dialogue generation tasks will increase, as AI-based evaluators can be be trusted to evaluate and support their performance in a comparable manner as human evaluators can do. There are two main types of evaluation: per-turn evaluation and per-dialog evaluation. We are focusing on the first one because, in a further work, we may use such evaluation on-line, measuring performances and using them to improve the running dialog. The evaluations we currently focus are related to the semantic meaning of the turn (in relation to the dialog), which requires an evaluation process able to extract such meaning. Our evaluation will be in-context evaluation, where the context is composed of a set of turn in a multi-party conversations, preceding the last one. For the human evaluation, we intend to use user-survey with multi-score questions and then automate that process by making a similar evaluation process where the human is substitute with an LLM-based evaluator.

Our study will contribute to the ongoing discourse on AI reliability and the future of human-AI interaction, ensuring that dialogues remain a trusted and integral part of our communication landscape. To this end, our work focuses on two primary objectives:

1. To measure how closely GPT-4o can approximate human-level performance in multi-party conversation quality assessments.

2. To measure how closely GPT-4o can approximate human-level performance in diadic dialogue turns error classification.

## 2 Background

### 2.1 Discussion Evaluation Methods

Evaluating dialogue systems has evolved significantly over the years, transitioning from manual human evaluations to sophisticated automated and machine learning-based methods. This section reviews the historical progression and current state of dialogue evaluation methodologies.

**Human-based Evaluation:** Initially, dialogue systems were primarily evaluated through human judgments. Human evaluators would assess dialogues based on various qualitative aspects such as coherence, relevance, informativeness, and engagement[13][22][26]. Metrics like *user satisfaction* and *task success rates* were common in early evaluations[17]. Human evaluations are considered the gold standard due to their ability to capture nuanced aspects of dialogue quality, but they are costly, time-consuming, and subject to inter-annotator variability and biases[41][13].

**Rule-based and Statistical Models:** With advancements in natural language processing, rule-based and statistical models were developed to automate dialogue evaluations. These methods often employed pre-defined rules or statistical measures to assess dialogue quality. For instance, metrics such as BLEU (Bilingual Evaluation Understudy) and ROUGE (Recall-Oriented Understudy for Gisting Evaluation) were initially used to evaluate the *overlap* between generated responses and reference texts [35][47]. However, these metrics often failed to capture the full quality of dialogues, as they focused on surface-level lexical similarities and did not account for the diversity of possible valid responses[47][29].

**Machine Learning Approaches:** The advent of machine learning brought more advanced methods for dialogue evaluation. Supervised learning models were trained on annotated datasets to predict human evaluation scores. Neural network models have also been explored for decision-making processes within dialogues, focusing on how rules and conditions are managed within a matrix-based system to enable complex decision-making tasks[33]. Notable models include ADEM (Automatic Dialogue Evaluation Model), which predicts *dialogue quality scores* based on human annotations[32]. Despite their improved performance over rule-based metrics, these models still required large amounts of annotated data for training and were limited by the quality and scope of the training data [43].

**Deep Learning and Neural Networks:** Recent years have seen a shift towards deep learning approaches, particularly with the introduction of pre-trained language models such as BERT and GPT. These models have been fine-tuned for dialogue evaluation tasks, leveraging large-scale, diverse datasets to better understand the context and nuances of dialogues[38][2]. For example, models like RUBER and its variants utilize both reference and unreferenced-based evaluation metrics, combining the strengths of both approaches [3].

**Evaluation Challenges and Multi-party Dialogue:** Evaluating multi-party dialogue systems presents unique challenges compared to dyadic systems (the vast majority of current systems). Multi-party dialogues involve complex interactions, turn-taking, and context management across multiple participants. Traditional metrics often fall short in capturing these dynamics. Researchers have proposed expanded taxonomies and evaluation frameworks to address these issues. For instance, the taxonomy by [24] includes specific error categories for multi-party dialogues, such as *speaker and addressee recognition errors*[34][51].

**Towards Standardized Evaluation Frameworks:** The field has recognized the need for standardized evaluation frameworks that can consistently assess dialogue systems across different tasks and contexts. Recent efforts include the development of comprehensive benchmarks and shared tasks, such as the Dialogue System Technology Challenge (DSTC) series, which provide standardized datasets and evaluation protocols for benchmarking dialogue systems[42]. These initiatives aim to facilitate the comparison of different systems and promote reproducibility and transparency in evaluation practices[8].

In summary, the evaluation of dialogue systems has progressed from relying solely on human-based methods to integrating automated metrics and machine learning models. Despite notable advancements in standardization and automation, continuous research and innovation are essential to overcome current challenges and improve the robustness and reliability of dialogue system evaluations. This paper propose a meta-analysis of the quality of purely LLM based evalu-ation methods compared to human evaluation that could allow to transition to fully automated conversation quality assessments.

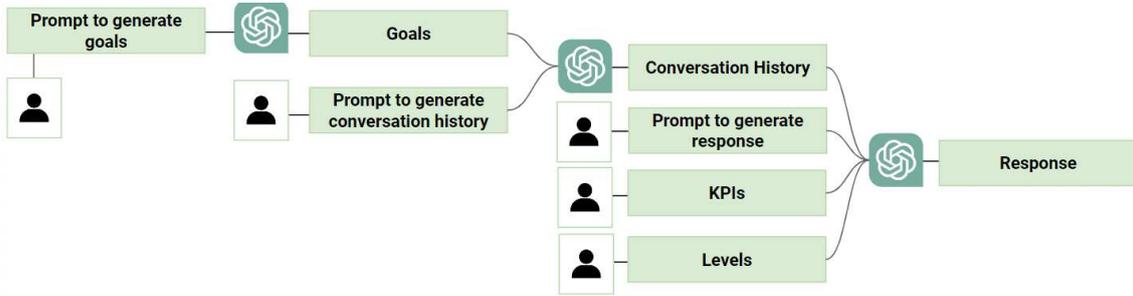

Figure 1: Architecture of Dataset creation for experiment 1.

## 2.2 Dataset

There are several dataset available providing dialog focused on different topics/domains [53]. The personal-Chat dataset [54] is highly focused on personal conversations between two individuals, making it very useful for studies on personalized dialogue agents. However, its narrow focus on personal conversation limits its applicability for multi-domain requirements. MultiWOZ dataset [7] is a comprehensive dataset for task-oriented dialogue modeling across multiple domains, which is crucial for many research purposes. Nonetheless, it is more focused on natural, open-domain dialogues rather than task-specific exchanges. [9], provides rich dialogue data from the TV show "Friends" although this corpus requires extensive processing to filter and select suitable dialogues. Similarly, the OpenSubtitles dataset [11], despite offering a large volume of movie dialogues, presented a similar challenge in terms of processing and selection of relevant data. Datasets like DREAM [46] was designed for dialogue-based reading comprehension. It primarily consists of descriptive text rather than conversational dialogue, which wasn't aligning with the objectives of this paper. Additionally the Entailment Bank dataset [12] is excellent for explanatory tasks and constructing entailment trees. However, it lacks the conversational structure necessary for dialogue-oriented study. ReDial [31] despite being a valuable resource for studying conversational recommendations, focus on movie recommendations and did not match our constraint for open-domain, multi-person conversations

Due to the limitations of existing datasets, a custom dataset was generated for this work, specifically designed to meet the requirements for multi-domain and multi-party dialogues without the constraints of extensive preprocessing or domain-specific limitations. Furthermore, an existing benchmark dataset was employed to evaluate dialogue error behaviors, offering a rigorous and contextually relevant assessment of system performance.

## 3 Experiments

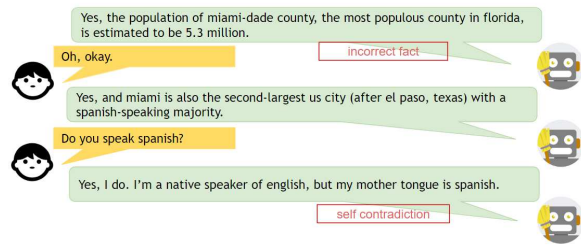

Figure 2: Example of the response errors in a human-bot dialogue for Experiment 2. Adapted from Finch et al. [18].

The experiments were organized into two parts. In the first experiment, to overcome the limitations of existing datasets, a carefully designed prompt was employed to generate a dataset using GPT-4o. This method enabled the construction of a dataset specifically tailored to the requirements of multi-party discussions and open dialogues. The need for a dataset that was clean, consistent, and aligned with the evaluation criteria, which existing datasets did not fully satisfy, was addressed through this approach. A total of 48 dialogues were compiled, each selected based on key performance indicators (KPIs) to ensure relevance to the study's objectives. The structure, as depicted in figure 1, was designed to align each discussion with the KPIs. By generating the dataset in-house, a high level of control over content, consistency, and relevance was maintained throughout the dataset.

> **GOAL:**
> Discuss your favorite movie and why?
>
> **DISCUSSION:**
>
> **Alice:** What's everyone's favorite movie and why?
>
> **Bob:** I love *"Inception"* because of its mind-bending plot and incredible visuals.
>
> **Carol:** My favorite is *"The Shawshank Redemption"* for its powerful storytelling and memorable characters.
>
> **Alice:** I adore *"Amélie"* because of its whimsical charm and uplifting message.
>
> **RESPONSE:**
>
> **John:** What's also concerning is how legislation will keep up with these rapid advancements to ensure they benefit society ethically.

Figure 3: In this Information-Seeking dialogue, the participants each respond directly to the question posed by Alice, providing personal insights and reasons for their choices. The focus is on gathering individual preferences and reasons, showcasing a straightforward exchange of information.

> **GOAL:**
> Discuss how cultural differences influence people's perspectives and interactions?
>
> **DISCUSSION:**
>
> **Maria:** Cultural differences shape how we interpret behavior, impacting workplace communication.
>
> **John:** Absolutely, cultural norms can dictate what is considered respectful or rude.
>
> **Liu:** This affects negotiations too, as expectations and negotiating styles differ across cultures.
>
> **Maria:** Agreed, understanding these differences can lead to more effective and harmonious interactions.
>
> **RESPONSE:**
>
> **John:** What's also concerning is how legislation will keep up with these rapid advancements to ensure they benefit society ethically.

Figure 4: In this Exploratory dialogue, Maria initiates the conversation by highlighting the impact of cultural differences on behavior interpretation and workplace communication. John builds on Maria's point by discussing how cultural norms influence perceptions of respect. Liu further explores the topic by mentioning how these differences affect negotiations. Maria then responds to Liu, reinforcing the importance of understanding cultural differences for better interactions.

In the second experiment, the ABC-Eval dataset from Finch et al. (2023) [18] was utilized as the benchmark for behavior error detection. This dataset comprises 400 open-domain human-bot dialogues, collected from interactions between university students and four different chatbots: BlenderBot2, BlenderBot with DECODE reranking, Emora, and Bart-FiDRAG. Each bot response in these dialogues was reviewed by human annotators, who identified the presence or absence of specific dialogue error behaviors. These annotations were carried out by crowdworkers on the SurgeHQ platform, who received training using three curated conversations to ensure accurate identification of each dialogue behavior prior to participating in the annotation project. For example, in Figure 2, the three bot responses are annotated as 1, 0, 0 for the behavior "incorrect fact" (!Fac) and as 0, 0, 1 for the behavior "self-contradiction" (!Sel). This dataset provided a robust benchmark for evaluating the system's performance across various conversational scenarios.

### 3.1 Experiment 1

#### 3.1.1 Designing Discussion Goals

In the first experiment, the creation of the dataset required well designed topics of discussion, referred to as *discussion goals*, which were structured as questions. Generating these goals involved a comprehensive and multi-step process. A specialized prompt was crafted to develop six topics of discussion, ensuring a range of complexity from straightforward to highly intricate. This process demanded careful consideration to balance relevance, engagement, and difficulty. Below is and example of a generated goal:

1. **Easy** - "Discuss your favourite movie and why?"

2. **Complex** - "Discuss the ethical implications of advancements like AI and biotechnology?"

This task required significant effort and precision to ensure that the topics were both diverse and challenging, enabling us to thoroughly evaluate the performance and capabilities of the dialogue system across various scenarios. The dataset creation process followed the structure depicted in Figure 1.

#### 3.1.2 Designing Discussions

After the *discussion goals* were established, each goal was used to generate dialogues between four participants. The discussions were predominantly categorized into *Information-Seeking* (IS) dialogues (see example in Figure 3) and *Exploratory dialogues* (see example in Figure 4. the former dialogues are structured primarily to acquire information from participants [49], while the later involve participants building upon the conversation initiated by a previous participant [27]. To generate these dialogues, prompts were designed to ensure that the conversations were relevant, coherent, and aligned with the evaluation criteria.

### 3.1.3 Designing Response Turn

The last step was to generate a *response turn* to all 6 discussions.

```
GOAL:
Discuss the ethical implications of advancements like AI and biotechnology?

DISCUSSION:
Samantha: How do you feel about the ethical implications of AI and biotechnology advancements?
Robert: I'm concerned that rapid AI development might lead to unforeseen societal disruptions and job displacement.
Linda: Biotechnology advancements raise ethical dilemmas, particularly regarding gene editing and the potential for designer babies.
Samantha: True, both fields necessitate careful consideration of privacy, consent, and long-term impacts on humanity.

RESPONSE:
Robert: Did you know that dolphins sometimes sleep with one eye open?
```

Figure 5: Example of the next turn response generated from the given goal, the discussion history, and a low score (0%) Coherence KPI response.

```
GOAL:
Discuss the ethical implications of advancements like AI and biotechnology?

DISCUSSION:
Samantha: How do you feel about the ethical implications of AI and biotechnology advancements?
Robert: I'm concerned that rapid AI development might lead to unforeseen societal disruptions and job displacement.
Linda: Biotechnology advancements raise ethical dilemmas, particularly regarding gene editing and the potential for designer babies.
Samantha: True, both fields necessitate careful consideration of privacy, consent, and long-term impacts on humanity.

RESPONSE:
John: What's also concerning is how legislation will keep up with these rapid advancements to ensure they benefit society ethically.
```

Figure 6: Example of the next turn response generated from the given goal, the discussion history, and a high score (100%) Coherence KPI response.

Every discussion has to have a different response which should be evaluated in relation to the dialog, the given KPI and a score level (see example in figure 5 and 6).

**KPI Definitions**: The selection of the four following key performance indicators (KPIs) was driven by the need to assess dialogue quality in a manner that reflects the nuances of human communication. Assessing the quality of dialogues is inherently subjective and varies based on text comprehension and judgment. Therefore, KPIs were chosen to capture essential aspects of effective communication that are not easily quantifiable but are crucial for evaluating dialogue systems. Our approach aligns with research emphasizing the importance of human-centric evaluation criteria in conversational AI [30][13]:

```
GENERATE RESPONSE:
Take the role of an English Language Professor.

Given the following conversation between various people:
<Goal/Question>

Knowing that the goal of the conversation is the following:
<goal>

Your task is to generate the next turn of this conversation, with a <kpi> level of <level>.

By <kpi>, I mean:
<kpi definition>

It is very important that your turn respects the level of <kpi> above, because it will be used to educate students.

Keep in mind that the beginning of the conversation may not have the same level of <kpi>.

Next Turn (In less than 20 words):
```

Figure 7: Example of a prompt used to query GPT-4o to generate a response: Note: every word enclosed in <>were replaced with the actual Goal/Question, KPI, Conversation, and level.

- **Coherence**: The degree to which the *response* relates to the discussion.

- **Innovation**: The degree to which the *response* adds new and creative ideas or solutions to the discussion, promoting novel approaches and perspectives

- **Concreteness**: The degree to which the *response* adds specific, tangible, and clear information or details to the discussion to enrich it."

- **Goal Contribution**: The degree to which the *response* adds relevant and effective points to the discussion, advancing towards the overall objective or outcome.

**Score Levels**: In addition to generating responses for each KPI, we also classified each response into different *score levels*. By "levels," we refer to specific percentages (0 - 100%) that indicate the degree to which a response meets the criteria for a given KPI. These *score levels* were chosen to have a 20% step to clearly reflect distinct variations in response quality. This granular-ity allows for a more nuanced evaluation of the responses, ensuring that subtle differences in quality are captured and can be analyzed effectively.

This structured approach ensures that the evaluation process is systematic and transparent, allowing for precise comparisons and analysis of dialogue quality across different KPIs and levels.

**Model Settings**: Table 1 contains the following settings for GPT-4o model during all experiments:

| Settings | Value |
|---|---|
| Temperature | 0.8 |
| Top-p | 1.0 |
| Max Tokens | 40% |
| Frequency Penalty | 0 |
| Presence Penalty | 0 |
| Stop Sequences | None |
| Logit Bias | None |

Table 1: Hyperparameter settings used for the model.

### 3.1.4 Methodology

After generating a suitable data set, a survey was created for humans and and an experiment "replicating" the survey was done requesting GPT to make the evaluation.

**Humans**: Four Google Forms were created, each focusing on a specific Key Performance Indicator (KPI). Initially, each form contained six goals, discussions, and corresponding responses. Subsequently, these forms were enhanced by incorporating a verification step to determine whether humans could reliably discern identical output levels in responses. This enhancement involved duplicating the responses based on different levels, ensuring that each goal and discussion included two responses graded at the same level. The objective was to evaluate whether humans could accurately detect when two sentences were rated at the same KPI level.

**GPT-4o**: A clear prompt was crafted that included goals, discussions, and individual responses to assess the KPI rating for each response. The number of responses was expanded from 6 to 12 to evaluate GPT-4o's reliability in identifying two sentences generated at the same KPI level and to compare its performance with that of humans.

### 3.1.5 Results

- **Human Responses:** We had a total of 33 participants, evenly distributed across all four forms, ensuring no overlap between participants. Each participant contributed unique responses, providing a diverse set of data for analysis.

- **GPT-4 Generated Responses:** For each human response, we utilized GPT-4 to generate corresponding outputs. This allowed us to calculate the average ratings and observe the variance in GPT-4's ratings for each question. In total, we generated 33 GPT-4 responses, covering all four Key Performance Indicators (KPIs). This approach enabled us to compare human and GPT-4 responses systematically, evaluating the consistency and accuracy of GPT-4 in replicating human-like responses.

### 3.1.6 Terminology

- **Human Rating**: Ratings assigned by human evaluators based on their opinion of the quality and relevance of each response.

- **GPT Rating**: Ratings assigned by GPT-4o based on its assessment of the quality and relevance of each response.

- **GPT Generation**: The predefined rating level used as a guideline for GPT4.o to generate the dataset responses.

### 3.1.7 Parameters

Both linear regression and sigmoid fitting techniques were used to analyze the data. As depicted in figure 9, the linear fit, represented by the green line, shows a strong correlation coefficient of 0.997, suggesting a strong linear relationship. However, the sigmoid function provides an even better fit, with a near-perfect correlation coefficient of 1.00. This improved fit is largely due to the sigmoid function's higher number of free parameters (three compared to two for the linear model), allowing it to capture more nuances in the data. The results highlight the binary nature of both human and GPT evaluations, where responses are often judged as either meeting or not meeting certain criteria, rather than on a continuous scale. This binary behavior aligns better with the sigmoid function, which effectively models such step-like changes in the data.

**Coherence KPI**: Figure 8a shows the similarity between human ratings and GPT ratings when

## Average Human vs. GPT Rating

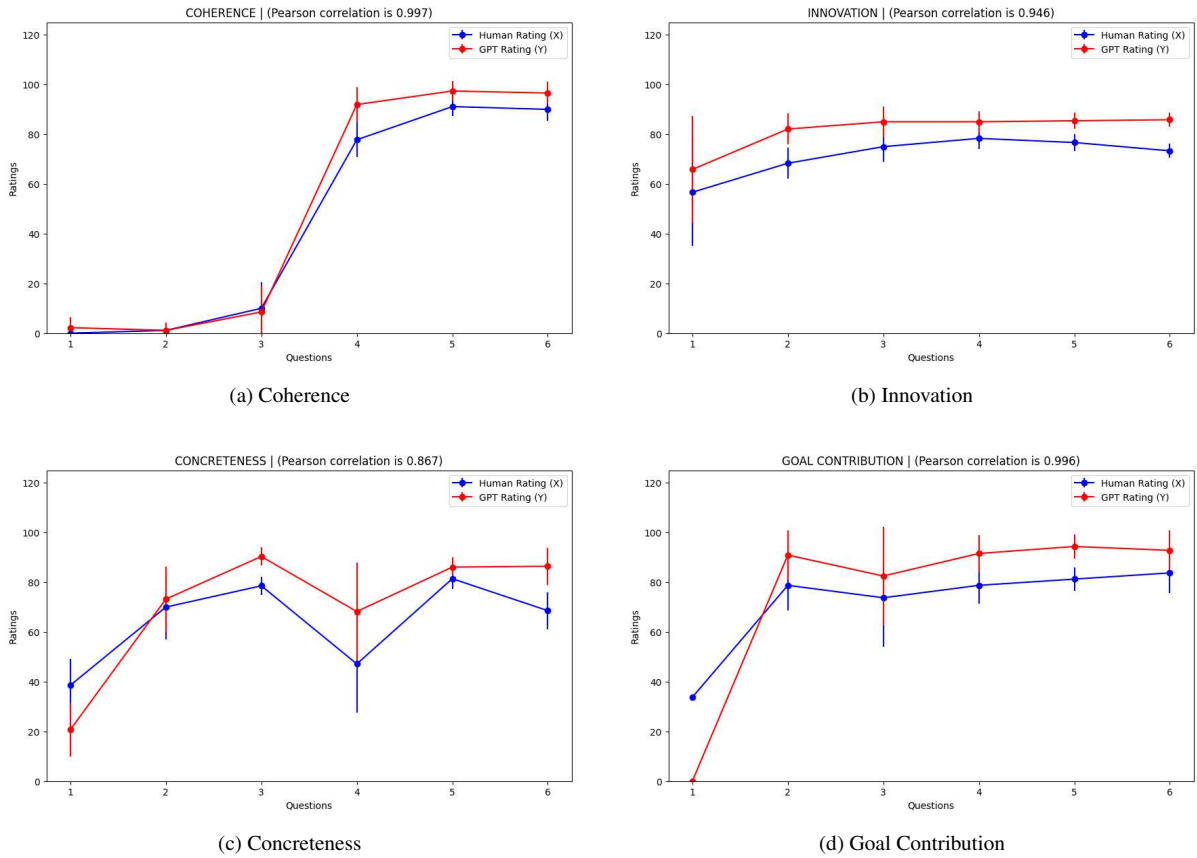

Figure 8: The chart displays the average ratings of humans and GPT across all four KPIs. Coherence has the highest correlation, although the response shows clearly a binary classification. Innovation seems to be very difficult to judge for both Human and LLM (the range of response is restricted to [50-90]). Concreteness has the lowest by correlation, between human and LLM nevertheless showing the same kind of response difficulty for both judges. Human and LLM shows also a binary behavior for goal contribution but with a bias towards high ratings.

## Comparative Analysis of Rating Correlations Between Human and GPT-4 Assessments

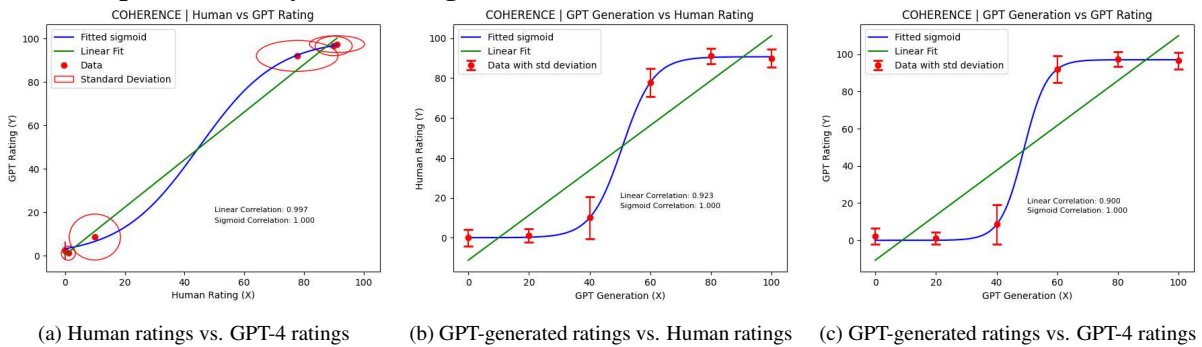

Figure 9: The plots illustrate the linear and sigmoid correlations between different rating sources. (a) shows the correlation between Human ratings and GPT-4 ratings, (b) shows the correlation between GPT-generated ratings and Human ratings, and (c) shows the correlation between GPT-generated ratings and GPT-4 ratings. The fitted models and standard deviations are indicated by red circles.

| KPI | Human vs. GPT Rating (Linear) | Human vs. GPT Rating (Sigmoid) | GPT Generation vs. Human Rating (Linear) | GPT Generation vs. Human Rating (Sigmoid) | GPT Generation vs. GPT Rating (Linear) | GPT Generation vs. GPT Rating (Sigmoid) |
|---|---|---|---|---|---|---|
| **COHERENCE** | 0.99 | 1.0 | 0.92 | 1.0 | 0.90 | 1.0 |
| **INNOVATION** | 0.95 | 0.99 | 0.75 | 0.97 | 0.75 | 1.0 |
| **CONCRETENESS** | 0.87 | 0.96 | 0.47 | 0.72 | 0.71 | 0.96 |
| **GOAL CONTRIBUTION** | 0.99 | 1.0 | 0.74 | 0.96 | 0.67 | 0.95 |

Table 2: The table shows the correlation values between different ratings, demonstrating that GPT ratings closely align with human ratings across various KPIs, with particularly high correlations for Coherence and Goal Contribution.

evaluated using the same Key Performance Indicator (KPI) definitions, with a correlation coefficient ($r$) of 0.99. This high correlation reflects a strong agreement between human evaluators and the GPT model's ratings. Figure 9 further explores the correlations at different levels generated by GPT. In these plots, standard deviations of the ratings are depicted using red ellipses, where the width and height of each ellipse correspond to the standard deviations in human ratings and GPT ratings, respectively. Both the fitted sigmoid and linear models are included to provide a visual comparison of the two fitting approaches. Table 2 presents all correlation values for each KPI. For coherence, the correlation coefficients between human ratings and GPT-generated ratings are 0.92 and 0.90, respectively, highlighting a significant alignment between human assessments and GPT's evaluations. Overall, the results indicate that human ratings and GPT ratings are more closely aligned with each other than the variations observed in the ratings generated by the GPT model itself. This alignment suggests a high degree of consistency and reliability in GPT's ratings, closely mirroring human judgment.

**Innovation KPI**: Analysis of the innovation responses reveals strong correlations between human ratings and GPT ratings, as well as between GPT-generated ratings and human ratings. Specifically:

- The linear correlation between human ratings and GPT ratings for each question is 0.95, while the sigmoid correlation is slightly higher at 0.98. This indicates a very high degree of agreement between human evaluators and the GPT model's initial ratings.

- The linear correlation between GPT-generated ratings and human ratings for innovation is 0.75, whereas the sigmoid correlation is a perfect 1.0. This suggests that while there is a moderate linear relationship, the nonlinear relationship as captured by the sigmoid model shows an ideal alignment.

- Similarly, the linear correlation between GPT-generated ratings and GPT ratings for innovation is also 0.75, with the sigmoid correlation being 0.97. This pattern further confirms that the nonlinear sigmoid model better captures the relationship between these ratings.

The correlation between the observed GPT ratings and the predicted ratings using the sigmoid function was 0.98, confirming the strong nonlinear relationship.

The linear correlation between the observed GPT ratings and the predicted ratings was 0.95, indicating a very high linear relationship.

Overall, these results demonstrate a significant alignment between human ratings and GPT ratings, with the sigmoid model capturing this relationship more accurately than the linear model. The sigmoid model's superior fit can be attributed to its ability to handle non-linearities and saturation effects in the data, which are often present in human judgment. This high level of agreement suggests that GPT-generated ratings are reliable and closely mirror human judgment in the context of innovation.

**Concreteness KPI**: For concreteness, specifically:

- The linear correlation between human ratings and GPT ratings for each question is 0.87, while the sigmoid correlation is higher at 0.96. This indicates a strong agreement between human evaluators and the GPT model's ratings, with the sigmoid model providing a more accurate reflection of the relationship.

- The linear correlation between GPT-generated ratings and human ratings is 0.47, suggesting a moderate linear relationship. However, the sigmoid correlation is significantly higher at 0.72, indicating that the sigmoid model captures the nonlinear relationship more effectively.

- The linear correlation between GPT-generated ratings and GPT ratings is 0.71, while the sigmoid correlation is 0.95. This further demonstrates that the sigmoid model is more adept at capturing the binary behavior in the ratings.

**Goal Contribution KPI**: Finally the analysis of the "Goal Contribution" KPI reveals a strong alignment between human ratings and GPT ratings. The linear correlation between human and GPT ratings is very high at 0.99, while the sigmoid correlation achieves a perfect 1.0. The correlation between GPT-generated ratings and human ratings is 0.74 with the linear model but improves significantly to 0.96 with the sigmoid model. Similarly, the correlation between GPT-generated ratings and GPT ratings is 0.67 for the linear model and 0.95 for the sigmoid model.

These findings suggest that GPT ratings are highly consistent with human ratings, especially when using the sigmoid model, which better captures the nuances in the data. Overall, GPT's ratings for "Goal Contribution" closely match human judgments, demonstrating yest again the model's reliability and accuracy.

### 3.1.8 Statistical Significance analysis

Statistical significance tests were conducted to compare human ratings, GPT ratings, and GPT-generated responses to ensure that the observed results were not attributable to random chance. The Shapiro-Wilk test was employed to determine if the ratings followed a normal distribution. Depending on the distribution results, either the paired t-test or the Wilcoxon Signed-Rank test was applied. For the comparison between human ratings and GPT ratings, the Shapiro-Wilk test indicated a normal distribution for coherence (p-value = 0.56), innovation (p-value = 0.83), and concreteness (p-value = 0.42). The paired t-test for coherence showed no significant difference

(p-value = 0.103), suggesting similarity between human and GPT ratings for coherence. In contrast, the paired t-test for innovation revealed a significant difference (p-value = 0.0002), indicating that GPT ratings for innovation differ significantly from human ratings. For concreteness, the paired t-test showed no significant difference (p-value = 0.29), indicating comparable ratings by humans and GPT. In the case of goal contribution, the Shapiro-Wilk test indicated a non-normal distribution (p-value = 0.0003), and the Wilcoxon Signed-Rank test showed no significant difference (p-value = 0.44), suggesting similar ratings by humans and GPT.

| HUMAN RATING VS GPT RATING | | Normally Distributed (Shapiro-Wilk Test) >0.05 | Paired t-test | | Wilcoxon Signed-Rank Test | |
|---|---|---|---|---|---|---|
| No. | KPI | Shapiro-Wilk Test Result | T-statistic | p-value (>0.05) | w-statistic | p-value (>0.05) |
| 1 | COHERENCE | Yes – 0.56 | -1.99 | 0.103 | | |
| 2 | INNOVATION | Yes – 0.83 | -9.59 | 0.0002 | | |
| 3 | CONCRETENESS | Yes – 0.42 | -1.19 | 0.29 | | |
| 4 | GOAL CONTRIBUTION | No – 0.0003 | | | 6.0 | 0.44 |

| HUMAN RATING VS GPT GENERATION | | Normally Distributed (Shapiro-Wilk Test) >0.05 | Paired t-test | | Wilcoxon Signed-Rank Test | |
|---|---|---|---|---|---|---|
| No. | KPI | Shapiro-Wilk Test Result | T-statistic | p-value (>0.05) | w-statistic | p-value (>0.05) |
| 1 | COHERENCE | Yes – 0.91 | -0.67 | 0.53 | | |
| 2 | INNOVATION | Yes – 0.75 | 1.64 | 0.16 | | |
| 3 | CONCRETENESS | Yes – 0.38 | 1.04 | 0.34 | | |
| 4 | GOAL CONTRIBUTION | Yes – 0.92 | 1.99 | 0.10 | | |

| GPT RATING VS GPT GENERATION | | Normally Distributed (Shapiro-Wilk Test) >0.05 | Paired t-test | | Wilcoxon Signed-Rank Test | |
|---|---|---|---|---|---|---|
| No. | KPI | Shapiro-Wilk Test Result | T-statistic | p-value (>0.05) | w-statistic | p-value (>0.05) |
| 1 | COHERENCE | Yes – 0.98 | -0.04 | 0.97 | | |
| 2 | INNOVATION | Yes – 0.62 | 2.42 | 0.60 | | |
| 3 | CONCRETENESS | Yes – 0.49 | 1.93 | 0.11 | | |
| 4 | GOAL CONTRIBUTION | Yes – 0.80 | 2.13 | 0.09 | | |

Table 3: Summary of significance testing results comparing human ratings, GPT ratings, and GPT-generated responses.

These statistical tests confirmed that the observed differences between human ratings, GPT ratings, and GPT-generated responses are consistent and not due to random variation. This enhances the reliability of the results and suggests that GPT-4's performance in these evaluations is robust.

## 3.2 Experiment 2

### 3.2.1 Overview

In addition to evaluating dialogue using arbitrary scores for each KPI, another method involves assessing whether system-generated utterances contain errors. This approach provides a straight-forward means of determining the quality of the system-generated responses.

In [18], error judgment results from GPT-3.5-turbo were compared with human judgment results as a method of evaluating a robot dialogue system. The objective of our experiment was to evaluate the error judgment results generated by the GPT-4.0-omni language model in comparison with the aforementioned results.

### 3.2.2 Key Performance Indicators

The accuracy of responses was evaluated based on the following KPIs shown in figure 10 below. In this section, of the KPIs indicated in [18], only those related to errors are dealt with.

| Type of KPI | Description |
| --- | --- |
| Commonsense contradiction | The response misunderstands or contradicts common knowledge. |
| Incorrect fact | The response hallucinates or inaccurately presents encyclopedic or expert knowledge. |
| Redundant | The response inappropriately repeats information presented earlier in the dialogue. |
| Ignore | The response ignores what the user just said. |
| Irrelevant | The response interrupts the current topic of discussion by presenting unrelated information. |
| Partner contradiction | The bot contradicts or misremembers something the user said earlier in the dialogue. |
| Self Contradiction | The bot contradicts something it said earlier in the dialogue. |

Figure 10: List of kpis.

### 3.2.3 Results

The dataset used in their study is publicly available and comprises 108 dialogue sets, including evaluations by two human judges and results from GPT-3.5-turbo. For our analysis, the same publicly available prompts were used and newly collected results using GPT-4.0-omni were generated. The analysis involved calculating the accuracy for each dialogue and then computing the average accuracy, which was subjected to a significance test. The mean accuracy was nearly identical to the accuracy calculated from the results of all dialogues, accurate to three decimal places.

When calculating the accuracy between human judges and each GPT model, if there was a discrepancy between the human judgments, the judgment that matched each GPT model's result was used to calculate the accuracy. Consequently, the accuracy for some KPIs may exceed that of the human-to-human comparisons.

The results of the comparison between GPT-4.0-omni, GPT-3.5-turbo, and human responses are illustrated in the accompanying bar chart. The chart presents the average accuracy and specific error rates for each key performance indicator (KPI).

- **Overall Accuracy (ALL):** The mean results across all items in all dialogues indicate that GPT-4.0-omni's accuracy is comparable to that of human responses. Statistical significance testing also revealed a significant difference between the performance of GPT-3.5-turbo and GPT-4.0-omni, demonstrating that GPT-4.0-omni is markedly more advanced than GPT-3.5-turbo and is effective in identifying errors in system utterances.

- **Incorrect Facts:** Human responses showed the highest accuracy in detecting incorrect facts, with statistically significant differences observed between human responses and those of both GPT models. Although GPT-3.5-turbo had slightly higher results than GPT-4.0-omni, the difference was not statistically significant. This suggests that while humans are more sensitive to incorrect information, GPT models can also be reliably used for this purpose.

- **Redundant Responses:** All three—GPT-4.0-omni, GPT-3.5-turbo, and human responses—exhibited similar accuracy in detecting redundant system utterances, with no significant differences observed among them. ANOVA results confirmed the lack of significant differences across the three groups.

- **Ignoring User Input:** When assessing whether the system ignored user utterances, GPT-4.0-omni outperformed GPT-3.5-turbo, though human responses still achieved the highest accuracy.

- **Irrelevant Information:** GPT-4.0-omni showed the highest accuracy in detecting irrelevant information, even surpassing human performance.

- **Commonsense Contradictions:** GPT-4.0-omni achieved the highest average accuracy in identifying commonsense contradictions, although the difference between its performance and that of GPT-3.5-turbo was not statistically significant.

- **Partner Contradictions:** In contrast, GPT-3.5-turbo showed the highest average accu-

racy in detecting partner contradictions, but again, there was no statistically significant difference between its performance and that of GPT-4.0-omni.

- **Self Contradictions:** Finally, for the self-contradiction KPI, human responses had the highest accuracy, but GPT-4.0-omni results were nearly equivalent, with no statistically significant difference between the two.

# 4 Conclusion

## 4.1 Experiment 1

The findings of Experiment 1 underscore the significant potential of GPT-based models in evaluating dialog through an automated process. Across various Key Performance Indicators (KPIs) such as Coherence, Innovation, Concreteness, and Goal Contribution, GPT ratings demonstrated strong alignment with human evaluations, particularly when analyzed through a sigmoid model. This alignment indicates that GPT models are capable of generating KPI evaluations that closely mimic human judgment, a critical aspect for their integration into real-world applications.

However, the variations observed between the linear and sigmoid correlations suggest that while GPT models are proficient in understanding and generating contextually appropriate responses, there remain nuances that are better captured through non-linear models. Moreover, the relatively lower correlations in the Innovation and Concreteness KPIs indicate areas where GPT models could benefit from further refinement. This discrepancy might be due to the subjective nature of terms like "Concreteness" and "Innovation," as human evaluators may interpret these concepts differently even when provided with specific definitions. Additionally, variations in prompt design could contribute to these differences, suggesting the need for further experiments to better understand and address these challenges.

## 4.2 Experiment 2

Experiment 2 provides valuable insights into the comparative accuracy of GPT-4.0 and human respondents in dialogue interactions. The detailed analysis of errors across various KPIs offers a clear understanding of where current AI models excel and where they need further refinement. This experiment contributes to the broader goal of developing more reliable and human-like conversational agents.

The experiment highlights the strengths and weaknesses of GPT-4.0 in generating accurate and coherent responses. While GPT-4.0 excels in maintaining factual accuracy and adhering to commonsense knowledge, it struggles with avoiding redundancy and self-contradiction. On the other hand, human respondents are generally better at maintaining coherence and avoiding self-contradiction but are more prone to commonsense contradictions and factual inaccuracies.

These findings suggest areas for improvement in future iterations of the GPT model, particularly in reducing redundancy and self-contradiction. Additionally, the results underscore the importance of continuous training and evaluation of language models to enhance their performance in dialogue systems.

# 5 Discussion

The collective insights from both experiments emphasize the strengths of GPT-4.0 in maintaining factual accuracy and commonsense knowledge while identifying areas for improvement such as redundancy and self-contradiction.

Moreover, the comparative analysis between human and GPT-4.0 responses highlights the complementary strengths of each. Humans excel in maintaining dialogue coherence and avoiding contradictions, while GPT-4.0 demonstrates superior factual consistency and commonsense understanding. This suggests that a hybrid approach, leveraging the strengths of both human judgment and AI capabilities, is the most effective strategy for developing advanced dialogue systems.

The observed differences in linear and sigmoid correlations in Experiment 1 indicate that non-linear models may better capture the nuanced aspects of human-like responses. This highlights the need for more sophisticated evaluation frameworks that can accommodate the complex nature of dialogue interactions.

In conclusion, these experiments contribute significantly to our understanding of the capabilities and limitations of GPT-based models in dialogue systems. The detailed analysis of errors and performance metrics provides a roadmap for future improvements, aiming towards more reliable and human-like conversational agents. The continuous evaluation and refinement of these models are cru-

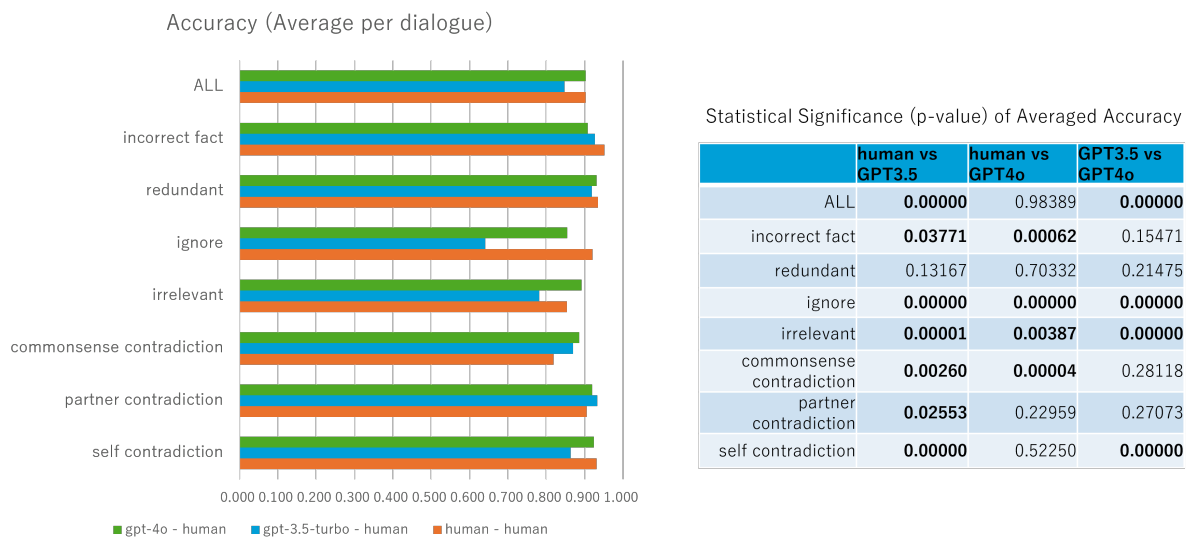

Figure 11: Accuracy result of experiment 2.

## 5.1 Challenges in Dialog Evaluation:

Despite significant progress, several challenges remain in the evaluation of dialogue systems. One major issue is the lack of correlation between automated metrics and human judgments. Research has shown that metrics like BLEU and ROUGE often do not align well with human perceptions of dialogue quality [20][42]. Additionally, there is a need for evaluation metrics that can handle the complexities of multi-party dialogues, where maintaining context and managing interactions between multiple participants add layers of complexity [20][51] . Future directions in dialogue system evaluation include developing more inclusive metrics that combine the strengths of human and automated evaluations. For example, prompting. Prompting has emerged as a powerful tool for dialogue evaluation by leveraging large language models [5]. It involves providing the model with specific instructions or context to generate and assess dialogue responses [55]. This method allows evaluators to simulate various conversational scenarios, measure the coherence, relevance, and informativeness of responses, and compare them to human judgments[1]. Prompting enhances the evaluation process by capturing nuanced aspects of dialogue quality and enabling automated, scalable assessments without the need for extensive annotated datasets[16]

## 5.2 Limitations and Future Works

Our experiment included human responses from a diverse range of individuals, from college students to university professors, which enriched the variety of opinions on each KPI definition. For future work, focusing on experts in the field of linguistics could provide more consistent evaluations, as these experts are likely to have more aligned interpretations of the defined terms.

Incorporating multi-modal inputs, such as visual or auditory data, could further enhance the model by providing a richer context, enabling it to generate more nuanced and contextually appropriate responses.

Expanding the range of KPIs in future studies would be a significant advancement in the field of NLP, particularly when working with a Black box like GPT.

## Acknowledgments

This work was partially supported by the Erasmus Mundus European Masters Program in Language and Communication Technologies (LCT).

# Annex

Figure 12: Sample of the questionnaire for human evaluation discussed in the methodology section

Figure 13: Example prompt to generate the questions, which serve as goals for our discussion

Figure 14: Guided definition of our KPI

Figure 15: This is an example prompt for generating discussions, using the role of an English Language Professor to guide the conversation.

17